\documentclass[sigconf]{acmart}
\usepackage{graphicx}
\usepackage{latexsym}
\usepackage{comment}
\usepackage{url}
\usepackage{booktabs}
\usepackage{float}
\newcommand\Alpha{\mathrm{A}}
\usepackage{enumitem}
\usepackage{amsfonts} 
\usepackage{amsmath}
\usepackage{tabularx}

\AtBeginDocument{%
  \providecommand\BibTeX{{%
    \normalfont B\kern-0.5em{\scshape i\kern-0.25em b}\kern-0.8em\TeX}}}

\acmPrice{15.00}

\newcommand{\cut}[1]{}

\newcommand{\aff}[1]{#1$^\alpha$}
\newcommand{\mca}[1]{#1$^m$}
\newcommand{\affmca}[1]{#1$^\chi$}

\newcommand{\task}{\texttt{ScienceWorld}}

\copyrightyear{2023}
\acmYear{2023}
\setcopyright{rightsretained}
\acmConference[K-CAP '23]{Knowledge Capture Conference 2023}{December 5--7, 2023}{Pensacola, FL, USA}
\acmBooktitle{Knowledge Capture Conference 2023 (K-CAP '23), December 5--7, 2023, Pensacola, FL, USA}
\acmDOI{10.1145/3587259.3627561}
\acmISBN{979-8-4007-0141-2/23/12}

\begin{document}

\title{Knowledge-enhanced Agents for Interactive Text Games}


\author{Prateek Chhikara}
\orcid{0000-0003-4833-474X}
\affiliation{%
  \institution{\small Information Sciences Institute, University of Southern California}
  \city{Los Angeles}
  % \state{California}
  \country{USA}
}
% \email{pchhikar@isi.edu}

\author{Jiarui Zhang}
\affiliation{%
  \institution{\small Information Sciences Institute, University of Southern California}
  \city{Los Angeles}
  % \state{California}
  \country{USA}
}

\author{Filip Ilievski}
\affiliation{%
 \institution{\small Information Sciences Institute, University of Southern California}
  \city{Los Angeles}
  \country{USA}
}

\author{Jonathan Francis}
\affiliation{%
 \institution{\small Human-Machine Collaboration, Bosch Center for Artificial Intelligence, USA}
 \country{}
 }

\author{Kaixin Ma}
\affiliation{%
  \institution{\small Language Technologies Institute, Carnegie Mellon University\\ Pittsburgh, USA}
  % \streetaddress{30 Shuangqing Rd}
  % \city{Haidian Qu}
  % \state{Beijing Shi}
  \country{}
  }

\renewcommand{\shortauthors}{Chhikara, et al.}

\begin{abstract}
Communication via natural language is a key aspect of machine intelligence, and it requires computational models to learn and reason about world concepts, with varying levels of supervision. Significant progress has been made on fully-supervised non-interactive tasks, such as question-answering and procedural text understanding. Yet, various sequential \textit{interactive} tasks, as in \cut{semi-Markov} text-based games, have revealed limitations of existing approaches in terms of coherence, contextual awareness, and their ability to learn effectively from the environment. In this paper, we propose a knowledge-injection framework for improved functional grounding of agents in text-based games. Specifically, we consider two forms of domain knowledge that we inject into learning-based agents: memory of previous correct actions and affordances of relevant objects in the environment. Our framework supports two representative model classes: reinforcement learning agents and language model agents. Furthermore, we devise multiple injection strategies for the above domain knowledge types and agent architectures, including injection via knowledge graphs and augmentation of the existing input encoding strategies. We experiment with four models on the 10 tasks in the \task~text-based game environment, to illustrate the impact of knowledge injection on various model configurations and challenging task settings.
Our findings provide crucial insights into the interplay between task properties, model architectures, and domain knowledge for interactive contexts.
\end{abstract}

%    <concept>
%        <concept_id>10010147.10010178.10010187</concept_id>
%        <concept_desc>Computing methodologies~Knowledge representation and reasoning</concept_desc>
%        <concept_significance>500</concept_significance>
%        </concept>
%  </ccs2012>

% \ccsdesc[500]{Computing methodologies~Knowledge representation and reasoning}


\keywords{Text-based Games, Knowledge Injection, Interactive Task Learning, Natural Language Communication}

\maketitle

\section{Introduction}
\begin{figure}[!h]
    \centering
    \includegraphics[width=\linewidth]{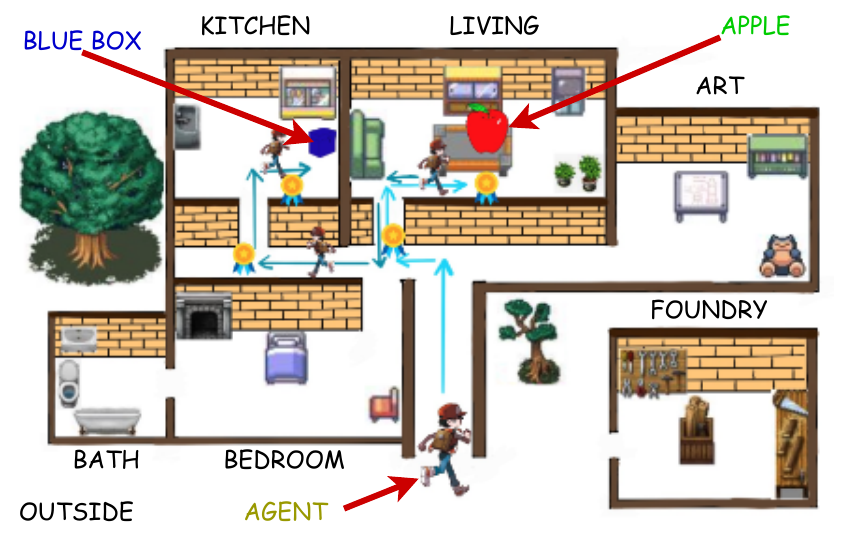}
    % \vspace{-0.5cm}
    \caption{Illustration of an Interactive Fiction (IF) game, where an agent must perform the task of picking a fruit (e.g., an apple) then placing it in a blue box in the kitchen.}
    \label{main}
    % \vspace{-0.4cm}
\end{figure}

Communication through natural language is a crucial aspect of machine intelligence \cite{s23}. The recent progress of computational language models (LMs) has enabled strong performance on tasks with limited interaction, like question-answering and procedural text understanding \cite{ma-etal-2022-coalescing,jiang2023transferring,s50}. Recognizing that interactivity is an essential aspect of communication, the community has turned its attention towards training and evaluating agents in interactive fiction (IF) environments, like text-based games, which provide a unique testing ground for investigating the reasoning abilities of LMs and the potential for AI agents to perform multi-step real-world tasks in a constrained environment. For instance, in Figure \ref{main}, an agent must pick a fruit in the living room and place it in a blue box in the kitchen. Text-based games use text instead of graphics, sounds, or animations to create interactive stories, and can include adventure, puzzle-solving, and role-playing themes; text-based games allow us to study models' abilities to perform functional grounding, separate from, e.g., the problem of multimodal grounding that is inherent in more-complex robot simulation environments \cite{francis2022core}. Recently developed text-based games, such as \textit{TextWorld} \cite{s24} and \textit{ScienceWorld} \cite{s1}, have quickly become popular, inspiring a variety of methods. To succeed in these games, agents must manage their knowledge, reason, and generate language-based actions that produce desired and predictable changes in the game world. 

IF games can be formulated as Partially Observable Markov Decision Processes (POMDPs), a category of sequential decision-making challenges under uncertainty. POMDPs encompass scenarios with only partially observable states and where the effects of actions are uncertain.
Thus, IF games can be modeled using reinforcement learning (RL)---with states, actions, observations, transitions, and rewards \cite{s25}. Observations correspond to text descriptions from environment and states are based on descriptions of agent and item locations, inventory contents, and surroundings. Given their natural language formulation, text-based games can also be tackled by LM approaches. The pros and cons of these two modeling paradigms are complementary. RL approaches function online and offer the advantage of modeling multistep transitions, but they can become challenging to optimize if reward structure and state information lack sufficient signals for effective learning. LMs offer flexibility in choosing subsequent actions, possess vast semantic knowledge, and can be advantageous for generating high-level, natural language instructions;
yet, they operate within rigorous constraints in input size and do not support multi-step interactions natively.

Prior work has shown that RL- and LM-based agents struggle to reason about or to explain science concepts in IF environments \cite{s1}, 
which raises questions about these models' ability to generalize to unseen situations beyond what has been observed during training
\cite{s27}. 
For example, while tasks such as `\textit{retrieving a known substance's melting (or boiling) point}' may be relatively simple, `\textit{determining an unknown substance's melting (or boiling) point in a specific environment}' can be challenging for these models. To improve generalization, it may be effective to incorporate world knowledge, e.g., about object affordances; yet, no prior work has investigated this direction.
In addition, existing models struggle to learn effectively from environmental feedback. For instance, when examining the conductivity of a specific substance, the agent must understand that it has already obtained the necessary wires and the particular substance so that it then proceeds to locate a power source.
Therefore, there is a need for a framework that can analyze and evaluate the effectiveness of different types of knowledge and knowledge-injection methods for text-based game agents.

In this paper, we design such a framework to augment existing text-based game agents with additional knowledge. 
We perform knowledge injection for two complementary paradigms based on training objectives: (1) online policy optimization through rewards, including \textit{pure} RL \cite{s2} and \textit{enhanced RL with Knowledge Graphs (KGs)} \cite{s3}, and (2) single-step offline prediction, including both
 \textit{pre-trained} LM \cite{s29} and \textit{instructions-tuned} \cite{lin2023swiftsage} LM.
We consider these two model classes because they are representative of the existing approaches for text-based games, which 
allows us to investigate how different model paradigms respond to knowledge injection techniques.
We experiment with two types of additional knowledge---namely, task history and object affordances knowledge. 
We evaluate the effectiveness of our proposed framework on the diverse set of 10 elementary school science tasks of the \task~environment \cite{s1}. 
The results illustrate that knowledge injection exerts a more favorable influence on single-step offline prediction models, i.e., LMs. Also, adding affordance knowledge is more beneficial than historical knowledge. Our contributions are as follows:

\begin{enumerate}[leftmargin=*]
\itemsep0em
    \item We investigate the role of knowledge injection in learning-based agents for semi-Markov interactive text games. We specifically focus on injecting memory about previous correct actions and the affordances of the relevant objects in the agent's scene.
    \item We integrate our injection strategies in two model paradigms, each with two variants: RL (`pure' RL and KG-enhanced RL) and language modeling (pre-trained and instructions-tuned). We devise multiple injection strategies to enrich the information --- as part of existing inputs, as new inputs, or as KG relations.
    \item We perform experiments on diverse tasks of \task~environment to provide insights on the impact of including affordance knowledge and action memory across different architectures, tasks, and knowledge-injection strategies. Our extensive experiments advance the understanding of how external knowledge can produce better action selection in text-based games.
\end{enumerate}

\section{Related Work}

\textbf{Reinforcement Learning for Text-based Games} has been a popular idea due to the conventional formulation of text-based games as Markov decision processes. 
A common challenge in these games is the combinatorially large action space, which makes it difficult to find a good policy. Carta \textit{et al.} \cite{s20} proposed an approach to achieve alignment through functional grounding, where an agent uses an LM as a policy to solve goals through online RL. Madotto \textit{et al.} \cite{s12} introduced a new exploration and imitation-based agent to play text-based games, which can be seen as a testbed for language understanding and generation. The proposed method uses the exploration approach of Go-Explore \cite{s13} for solving games and trains a policy to imitate trajectories with high rewards. eXploit-Then-eXplore (XTX)~\cite{s17} is a multi-stage episodic control algorithm that separates exploitation and exploration into distinct policies, guiding the agent's action selection at different phases within a single episode. Yao \textit{et al.} \cite{s14} proposed a Contextual Action LM to generate a compact set of action candidates at each game state and combine it with an RL agent to re-rank the generated action candidates. The Deep Reinforcement Relevance Network (DRRN) model ~\cite{s2} uses a separate Gated Recurrent Unit (GRU) for processing action text into a vector which is used to estimate a joint Q-Value \textit{Q(o, a)} over the observation $o$ and each action $a$. Our work injects knowledge into the DRRN model to enhance agents' understanding of the game world. 
While these works have typically relied on the memory of the single previous action taken, regardless of its utility, our approach distinguishes itself by taking into account the memory of \textit{all} previous actions that generated a positive reward. Thus, our agents obtain better performance by using this information to reinforce correct decision-making and avoid repeating past mistakes. 

\noindent \textbf{LMs for Text-based Games} used in works such as Swift \cite{lin2023swiftsage}, ReAct \cite{yao2022react}, and SayCan \cite{brohan2023can} have revealed the feasibility of autonomous decision-making agents. 
Swift~\cite{lin2023swiftsage} is a model that takes into account the environment state and the history of the last ten actions as input strings for the LM. Additionally, they pursued model training through a supervised approach. 
ReAct~\cite{yao2022react} enables LMs to generate subgoals within action planning by incorporating a virtual \textit{`think'} action. This method necessitates human annotators to furnish instances of \textit{`thinking,'} outlining subsequent subgoals and furnishing comprehensive action trajectories. 
SayCan~\cite{brohan2023can} integrates an LM and a value function, aligning with grounding affordances. Using historical and current context as textual inputs, SayCan generates a ranked list of actions, grounding LMs through value functions reflecting action success likelihood. 
While Swift and SayCan retain a record of action history, the contribution of this information is not systematically studied. Moreover, they do not include world knowledge like object affordances.

\noindent\textbf{Knowledge-injection in Text-based Game Agents} has been used to enhance the performance of RL and LM agents. 
Ahn \textit{et al.} \cite{brohan2023can} identify potential actions using an LM and assign scores to these actions based on their likelihood of success in a given environment, which can be seen as an implicit affordance information. 
Swift \cite{lin2023swiftsage} introduced an additional layer of knowledge by incorporating the history of the previous ten actions within the episode.
Several works \cite{s11,s3,s15,s31} have used KGs as an extra knowledge source to provide a structured representation of the game world, which can be used to guide agents' decision-making. 
Xu \textit{et al.} \cite{s11} proposed a hierarchical framework built upon a KG-based RL agent to address generalization issues in text-based games; they achieve favorable results in experiments with various difficulty levels. KG-A2C \cite{s3} is a text-based game agent that builds a dynamic KG while exploring and generating actions. 
KG-DQN~\cite{s15} is a KG-based approach for state representation in deep RL agents, which involves building a KG during exploration and utilizing question-answering techniques to pre-train a deep Q-network for action selection. The authors in \cite{s31} formulated a state abstraction for common sense games by utilizing a subclass relationship from an open-source KG. Complementary to these methods, we provide a generic framework that supports various methods to inject explicit historical and affordance knowledge into text-based game agents, improving their effectiveness beyond previous KG approaches. 

\section{Framework for Knowledge Injection in Text-based Game Agents}

In most text-based games, the agent's input is comprised of three primary elements: the observation of the environment \textit{(obv)}, the contents of the agent's inventory \textit{(inv)}, and the task description \textit{(desc)}. These elements give the agent the context to make informed decisions and progress through the game. Based on these inputs, the agent is presented with a set of valid actions that it can perform, such as moving to a new location, interacting with objects in the environment, or using items in its inventory. Through these interactions, agents explore the game world, solve puzzles, and advance the story. In this section, we detail approaches for improving agents' downstream performance, thereby improving agents' coherence in action generation, their contextual awareness, and their abilities to learn effectively from the interactive environment. We consider two types of knowledge for enriching the inputs and two representative model classes as subjects for knowledge injection with their corresponding variants and knowledge-injection strategies.

\subsection{Input Enrichment with Extra Knowledge}

We expect that the raw inputs from the environment (observation, inventory, and task description) make it challenging for the agent to act coherently and learn from its mistakes. To improve the coherence and learning process, we enrich the apparent inputs with two complementary knowledge types: action memory and affordances. 

\textbf{Action Memory.} Historical knowledge is necessary for an AI agent to learn how to predict future steps based on a sequence of steps that it has taken previously. The historical knowledge could be in the form of all past actions picked by the model or the sequence of correct actions chosen by the model. 
Our analysis shows that preserving the past correct actions is a superior approach because it helps to reinforce successful strategies and prevent the model from repeating unsuccessful ones. Hence, we preserve the \textit{memory of previous correct actions} (MCA) taken by the agent in the current episode as input for all our models. 
Moreover, the memory can be short-term (within an episode) or long-term (across episodes). We focus on short-term memory from the current episode.
MCA is determined by the environment feedback. If an action yields a reward, then it is considered correct. Therefore correct actions cannot be fed to the agent initially, but are instead stored in memory as the agent progresses through the (train/test time) episode. 

\textbf{Affordance Knowledge.} 
Essentially, affordances are the set of possible actions allowed in a particular state of the environment.
Within the field of perceptual psychology, they are seen as a central tool through which living beings categorize their environment~\cite{gibson1977theory}.  
% Within linguistics, it has been shown that affordances play a central role in understanding the meaning of sentences \cite{Attardo:2005, Kaschak:Glenberg:2000}. 
We expect that affordances can help models learn better by listing the possible interactions with the objects around them. 
Unlike historical knowledge, the environment does not provide the affordances, and they need to be retrieved from external sources. For this purpose, we use ConceptNet \cite{speer2017conceptnet} and obtain its \textit{capableOf} and \textit{usedFor} relations for the objects in a given IF game episode.\footnote{\url{https://github.com/ease-crc/ease\_lexical\_resources}} The obtained affordances are then aggregated with the original environment inputs. For the example in Figure~\ref{main}, we inject information that an apple affords being eaten, and a box can contain objects.

\subsection{Knowledge Injection in Methods}
\label{subsec:arch}
We support two complementary paradigms based on training objectives: (1) online policy optimization through rewards using reinforcement learning (RL), where we frame the task as a POMDP; and (2) single-step offline prediction achieved through supervised training, approached as a language modeling task.

\begin{figure}[!t]
    \centering
    \includegraphics[width=\linewidth]{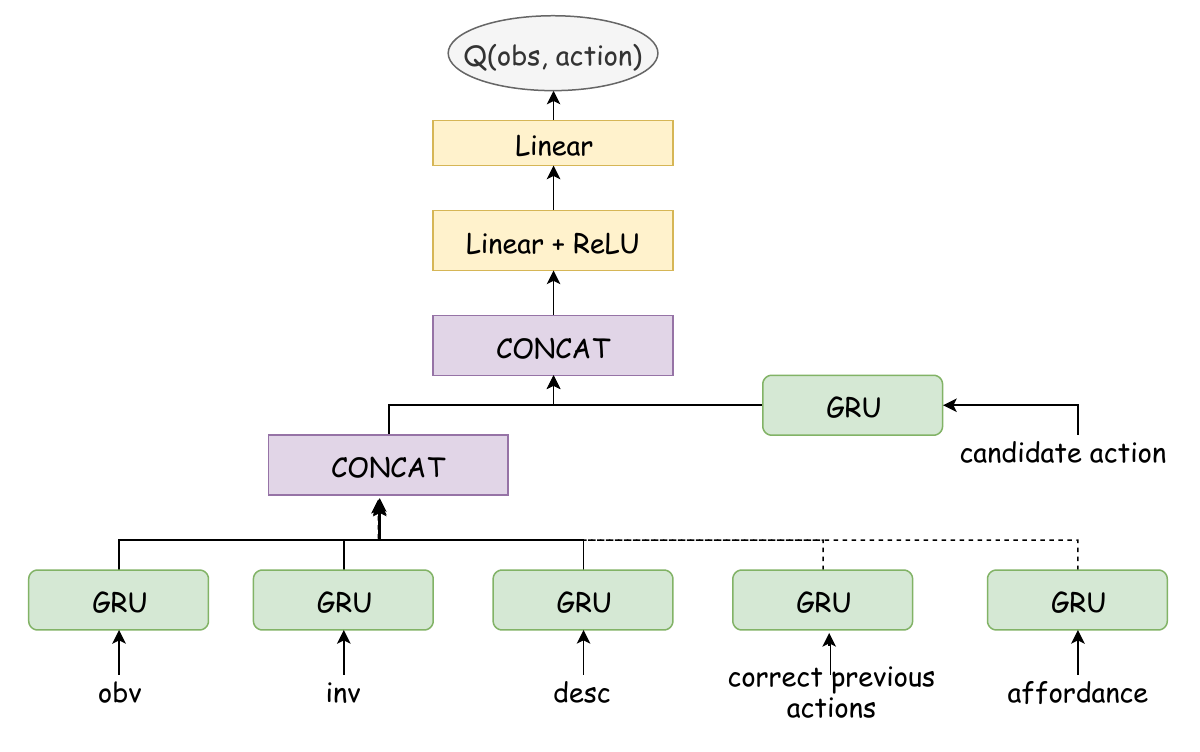}
    % \vspace{-0.5cm}
    \caption{DRRN architecture, enhanced with the memory of previous correct actions and object affordances.}
    % \vspace{-0.4cm}
    \label{drrn}
\end{figure}

\begin{figure*}[!t]
    \centering
    \includegraphics[width=0.6\linewidth]{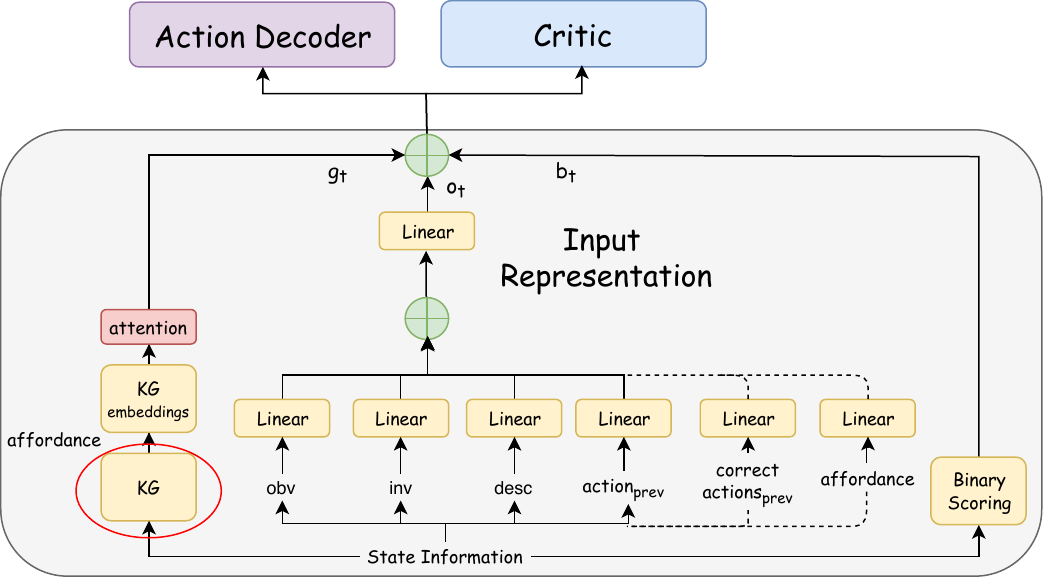}
    % \vspace{-0.2cm}
    \caption{KG-A2C model architecture with integrated affordances and previous correct actions.}
    \label{kga2c}
    % \vspace{-0.4cm}
\end{figure*}

\subsubsection{Online Policy Optimization through Rewards (RL Methods)}

\paragraph{\textbf{(1) Pure RL-based Model}}
We employ 
DRRN~\cite{s30}, due to its strong performance across challenging interactive text-based environments \cite{s2}. DRRN leverages a GRU to encode the current game state into a vector as shown in Figure \ref{drrn}. It uses a separate GRU to encode each of the valid actions into a vector and then combines the action vector with the game state vector through an interaction function to compute the Q-value (\texttt{Q}$_{a_i}$), which estimates the total discounted reward expected if that action is taken. The policy $\pi_{DRRN}$ is learned by maximizing the expected cumulative reward over time using the Q-values computed by the network. To stabilize training, DRRN uses experience replay and target networks.

\noindent \textbf{Knowledge Injection.} The \textbf{\texttt{baseline}} DRRN model uses only the inputs of observation, inventory, and task description to compute Q-values for each action.
To enhance the DRRN baseline, we have injected external knowledge into the model and created three new variations of DRRN:
    \textbf{\texttt{1.aff:}} Using a distinct GRU encoding layer, we introduce the affordances of the objects presented in the inputs to the baseline model in this approach.
    \textbf{\texttt{2.mca:}} A separate GRU encoding layer is utilized in this model to pass all previously correct actions to the baseline model.
    \texttt{\textbf{3.aff} $\oplus$ \textbf{mca:}} The encoding of this architecture is comprised of both the agent's previous correct actions and the affordance as distinct components:%
\begin{align}
{S}_{info} &\leftarrow {G}_{obv} \odot {G}_{desc} \odot {G}_{inv} \odot {G}_{aff} \odot {G}_{mca}\\
{Q}_{a_i} &\leftarrow \texttt{W}_{2}.(\texttt{ReLU}(\texttt{W}_{1}.({S}_{info} \odot {G}_{a_i}))),
\end{align}
where $\odot$ signifies concatenation, $G$ is GRU encoder of size F$\times$F, \texttt{W}$_{1}\in\mathbb{R}^{6F\times F}$, \texttt{W}$_{2}\in\mathbb{R}^{F\times 1}$, F is the embedding dimension of size 128, and ${Q}_{a_i}$ is the Q-value for each valid action $a_i$.

\paragraph{\textbf{(2) RL-enhanced KG Model:}}
As a knowledge-augmented RL agent, we used the Knowledge-augmented Actor-Critic (KG-A2C) model \cite{s3}. For KG-A2C, in addition to the textual representation of the game state, the agent also builds a dynamic KG representing the state space by parsing the textual descriptions using OpenIE \cite{angeli-etal-2015-leveraging}. KG's symbolic representation of the game states can help effective reasoning about the next course of action. The overall model architecture is shown in Figure \ref{kga2c}. Each of the textual inputs is encoded with a GRU, and the KG is separately encoded with KG embeddings and Graph Attention network \cite{s4}. In addition, the model takes into account the total score obtained so far through the binary score encoding. Formally, KG-A2C produces KG encoding ($g_t$), input encoding ($o_t$), and score encoding ($b_t$): 
\begin{align}
g_t &\leftarrow f(W \left(\bigoplus\limits_{k=1}^{K} \sigma (\sum_{j \in N} \Alpha^{k}_{ij}\cdot W^{k}\cdot h_j) \right) + b)\\
{o}_{t} &\leftarrow {G}_{obv} \odot {G}_{desc} \odot {G}_{inv} \\
{S}_{info} & \leftarrow g_t \odot o_t \odot b_t
\end{align}
where $\odot$ signifies concatenation, $G$ is GRU encoder of size 100$\times$100. \textit{W} and \textit{b} are weights and biases, $A_{ij}$ is the attention weights, and $h_j$ is the node feature vector. $b_t$ is the binary score encoding of the total score obtained so far with a shape of 1x10, which is calculated using the cumulative reward attained up to the present moment. The reward is first converted to a binary format with a length of 9, to which a `0' is added to the beginning in case the cumulative reward is positive, and a `1' is added if the reward is negative. The final state info vector ${S}_{info}$ is calculated by concatenating the three input representations, and it is then used to generate actions for the agent.
Overall, the model is trained with the actor-critic policy gradient. Instead of sampling directly from the valid action space, the policy network generates action templates and then populates the templates with objects from the knowledge graph. Thus, to make the learning more effective, KG-A2C also adds three auxiliary losses to encourage the model to generate valid actions, i.e., actions that would cause the game state to change:
\begin{align}
    \begin{split}
    \mathrm{L_T}(s_t, a_t; \theta_t) &\leftarrow \frac{1}{N} \sum_{i=1}^N (y_{\tau_i} \log \pi_{\mathbf{T}}(\tau_i|\mathbf{s}_t)\;\; + \\  
    & (1-y_{\tau_i})(1 - \log \pi_{\mathbf{T}}(\tau_i|\mathbf{s}_t))))
    \end{split}
    \\[1ex]
    \begin{split}
    \mathrm{L_O}(s_t, a_t; \theta_t)  &\leftarrow \sum_{j=1}^{n} \frac{1}{M} \sum_{i=1}^{M} (y_{o_i} \log \pi_{o_j} (o_i | \mathbf{s}_t)\;\; + \\  
    & (1 - y_{o_i}) (1 - \log \pi_{o_j} (o_i|\mathbf{s}_t))))
    \end{split} 
    \\[1ex]
    \begin{split}
    \mathrm{L_E}(s_t, a_t; \theta_t) &\leftarrow \sum_{a \in \mathcal{V}(s_t)} P(a|s_t)\log P(a|s_t) 
    \end{split} 
\end{align}
where $\mathrm{L_T}$, $\mathrm{L_O}$, and $\mathrm{L_E}$ are template loss, object loss, and entropy loss respectively. $a$ $\in$ Valid($s_t$) is a valid action, $\tau$ $\in$ Valid$_\tau$($s_t$) is valid template, $o$ $\in$ Valid($o_t$) is a valid object, and $s$ is a state.

\noindent \textbf{Knowledge Injection.} 
As \textbf{\textbf{\texttt{baseline}}}, we use a modified version of KG-A2C, where we utilize a single golden action sequence provided by the environment as the target, even though there may exist multiple possible golden sequences. We found this target to perform better than the original target of predicting a valid action. % in practice.
We devise the following knowledge-injection strategies to incorporate memory of correct actions and affordance knowledge for KG-A2C:  
\textbf{\textbf{\texttt{1. mca:}}} On top of the baseline, we incorporate all previously correct actions by using a separate GRU encoding layer and concatenate the output vector along with other output representations.
\textbf{\textbf{\texttt{2. aff:}}} The KG component in the KG-A2C model provides us with a convenient way to add more knowledge. In particular, we directly add the affordance knowledge into the KG as additional triples on top of the \texttt{baseline} model. For example, given the existing relation in the KG \textit{(living room, hasA, apple)} we can add the affordance relation: \textit{(apple, usedFor, eating)}. In this way, the KG encoding network can produce a more meaningful representation of the game state and potentially guide the model to produce better actions. In our experiments, we compare this approach to adding affordance knowledge using a separate GRU encoding layer, similar to the DRRN case. 
\texttt{\textbf{3.aff} $\oplus$ \textbf{mca:}} We include both affordances in the KG and the memory of all previous correction actions with a separate GRU encoding layer.

\begin{figure}[!t]
    \centering
    \includegraphics[width=\linewidth]{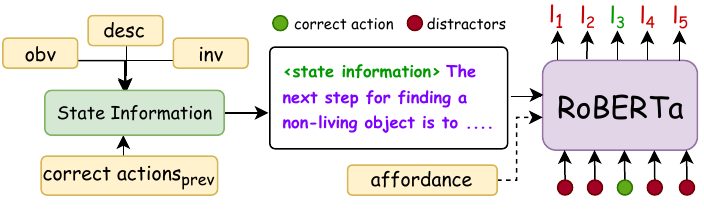}
    % \vspace{-0.3cm}
    \caption{RoBERTa architecture trained using distractors.}
    % \vspace{-0.4cm}
    \label{roberta}
\end{figure}

\subsubsection{Single-step Offline Prediction (LM Methods)}

\paragraph{\textbf{(1) Pre-trained LM:}}
We employed the RoBERTa \cite{s29} pre-trained LM due to its strong performance on various procedural understanding and commonsense reasoning tasks \cite{jiang2023transferring}.
RoBERTa is a transformer-based, encoder-only model trained using masked language modeling. Due to its large size, we choose offline fine-tuning to train the agent. Here we view the task as multiple-choice QA. At each step, the current game state is treated as the question and must predict the next action from a set of candidates. Similar to RL agents, the model is given the environment observation ($obv_i$), inventory ($inv_i$), and task description ($desc$) at every step. Then we concatenate it with each action and let the LM select the action with the highest score. 
Given the large set of possible actions, we only randomly select $n$ distractor actions during training to reduce the computational burden, the LM is trained with cross-entropy loss to select the correct action. 
At inference time, the model assigns scores for all valid actions, and we use top-p sampling for action selection to prevent it from being stuck in an action loop. 

\noindent \textbf{Knowledge Injection.} %To explore the best approach for enhancing RoBERTa's performance, 
We formalize three knowledge-injection strategies for the baseline RoBERTa model (Figure \ref{roberta}): 
\textbf{\textbf{\texttt{1.mca:}}} Here, we enable the LM to be aware of its past correct actions by incorporating an MCA that lists them as a string, appended to the original input. Due to token limitations of RoBERTa, we use a sliding window with size $A$, i.e., at each step, the model sees at most the past $A$ correct actions.
\textbf{\textbf{\texttt{2.aff:}}}
We inject affordance knowledge into the LM by first adapting it on a subset of the Commonsense Knowledge Graph~\cite{ilievski2021cskg} containing object utilities~\cite{s33}. We adapt the model via an auxiliary QA task following prior knowledge injection work~\cite{s32}.
We use pretraining instead of simple concatenation for input due to the substantial volume of affordance knowledge triples, which cannot be simply concatenated to the input of RoBERTa due to limited input length.
Pre-training on affordances through an auxiliary QA task alleviates this challenge, while still enabling the model to learn the relevant knowledge.
We then finetune our task model on top of the utility-enhanced model, as described in the baseline.
\textbf{\textbf{\texttt{3.aff $\oplus$ mca:}}} This variation simply combines \textbf{\texttt{mca}} and \textbf{\texttt{aff}}.

\begin{figure}[!t]
    \centering
    \includegraphics[width=\linewidth]{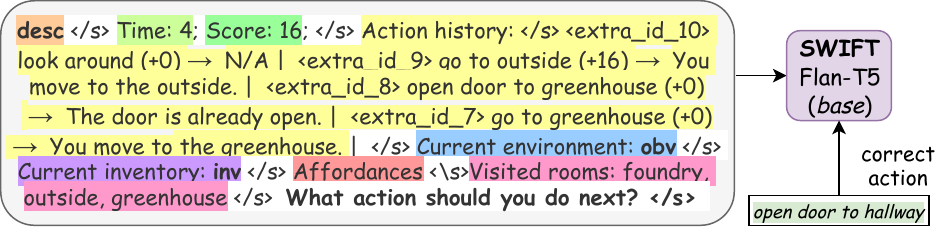}
    % \vspace{-0.3cm}
    \caption{Swift architecture trained in a Seq2Seq manner.}
    % \vspace{-0.5cm}
    \label{swift}
\end{figure}

\paragraph{\textbf{(2) Instruction-tuned LM:}}
We utilized the Swift model \cite{lin2023swiftsage}, which is based on the \texttt{Flan-T5} \cite{chung2022scaling} instruction-following architecture. The training follows a Seq2Seq methodology, wherein the input comprises state information, and the desired outcome is the correct action. The encompassed state information integrates task and environmental data: \texttt{``\textit{desc} – step number – score – action history – \textit{obv} – \textit{inv} – visited rooms – What action should you do next?"} (Figure \ref{swift}). The action history contains the last ten performed actions, each with the respective environmental reward, e.g., \texttt{``go to outside (+16) --> You move to the outside."}

\noindent \textbf{Knowledge Injection.}
The Swift model inherently integrates the historical context of the preceding ten actions. Notably, in contrast to the three previously examined models that exclusively consider the history of the last ten correct actions, the Swift model adheres to its original design by encompassing the entire history of the ten previous actions.
To establish a comparable \textbf{\texttt{baseline}} model to the methodology applied in the preceding three architectures, we omit the action history from the Swift model.
The unaltered variation of Swift is herein denoted as the \textbf{\texttt{1.mca}} version. Additionally, incorporation of affordance into the baseline model yields the \textbf{\texttt{2.aff}} model. Similarly, integration of affordances within the mca version led to the formation of the \textbf{\texttt{3.aff $\oplus$ mca:}} model. These affordances are introduced into the primary input sequence immediately following the inventory data and preceding information about visited rooms.

\begin{table}
  \centering
  \small
  \caption{Comparison of \texttt{baseline (base)} versus our best model configuration (\aff{Affordance}, \mca{MCA}, and \affmca{Affordance ~$\oplus$ MCA}) from \S\ref{subsec:arch}, based on average cumulative reward across task variants. \underline{Underlined} numbers indicate \textit{no} performance improvement over baseline. \textbf{Bold} numbers indicate the overall best model configuration for the task.}
  \label{tab:best}
   % \vspace{-0.3cm}
   \setlength{\tabcolsep}{3.5pt}
  % \resizebox{\columnwidth}{!}{
  \centering
  % \vspace{-0.6cm}
  \begin{tabular}{crrrrrrrr}
    \toprule
    & \multicolumn{2}{c}{\textbf{DRRN}} & \multicolumn{2}{c}{\textbf{KG-A2C}} & \multicolumn{2}{c}{\textbf{RoBERTa}} & \multicolumn{2}{c}{\textbf{Swift}} \\
    \midrule
    \textbf{Task} & \textbf{base} & \textbf{best} & \textbf{base} & \textbf{best} & \textbf{base} & \textbf{best}  & \textbf{base} & \textbf{best} \\
    \hline
    1 (2) & \textbf{04.82} & \underline{\aff{\textbf{04.82}}} & 01.48 & \affmca{03.70} & 00.74 & \affmca{01.48} & 00.00 & \mca{02.22} \\
    % \hline
    2 (4) & 07.07 & \aff{07.15} & 05.24 & \mca{13.59} & 01.48 &  \aff{\textbf{19.48}} & 07.57 & \aff{07.58} \\
    % \hline
    3 (8) & 07.67 & \aff{09.67} & 07.67 & \aff{08.33} & 02.67  & \mca{06.67} & 14.60 & \affmca{\textbf{51.60}}  \\
    % \hline
    4 (13) & 69.36 & \aff{70.15} & 70.96 & \mca{73.15} & 67.89 & \aff{71.11} & 86.36 & \aff{\textbf{99.00}} \\
    % \hline
    5 (16) & 08.89 & \affmca{09.71} & 05.92 & \aff{07.15} & 08.87 & \aff{\textbf{11.48}} & 06.39 & \mca{09.70} \\
    % \hline
    6 (17) & 20.34 & \mca{22.32} & 22.82 & \aff{23.07} & 07.79 & \underline{\affmca{07.39}} & 22.88 & \affmca{\textbf{25.00}}  \\
    % \hline
    7 (21) & 30.04  & \aff{30.73} & 31.42 & \affmca{32.29} & 14.41 & \aff{45.31} & 61.31 & \affmca{\textbf{89.63}} \\
    % \hline
    8 (23) & 09.87 & \affmca{14.07} & 15.93 & \affmca{17.53} & 04.00 & \underline{\mca{02.80}} & \textbf{45.20} & \underline{\aff{40.20}}  \\
    % \hline
    9 (27) & 09.37 & \affmca{09.92} & 07.78 & \mca{08.17} & 02.70 & \mca{04.68}  & 17.14 & \aff{\textbf{17.50}} \\
    % \hline
    10 (29) & 13.28 & \aff{16.94} & 12.93 & \underline{\aff{11.48}} & 03.77 & \affmca{04.67} & \textbf{17.16} & \underline{\mca{\textbf{17.16}}} \\ \midrule
    \textit{avg.} & 18.08 & 19.55 & 18.21 & 20.36 & 11.43 & 17.50 & 27.86 & \textbf{35.96}  \\
    \bottomrule
  \end{tabular}
  % }
   % \vspace{-0.5cm}
\end{table}

\begin{table*}
  \centering
  \small
  \caption{Baselines (\textbf{$b$}) comparison with knowledge-injected model configurations (``\texttt{$\alpha$}'': Affordance; ``\texttt{m}'': Memory of Correct Actions), based on average cumulative reward across task variants. \textbf{Bold} signifies better performance over baseline. %\textbf{Bold} numbers indicate the best performance for a specific model configuration and task. \underline{Underlined} numbers indicate no performance increment over baseline. 
  % \prateek[]{\perf{n} denotes the number of times the agent reached the perfect score of 100.} 
  }\label{tab:affordance_pca}% 
  % \vspace{-0.4cm}
  % \resizebox{\textwidth}{!}{
  \centering
  \begin{tabular}{c|rrrr|rrrr|rrrr|rrrr}
    \toprule
     & \multicolumn{4}{c|}{\textbf{DRRN}} & \multicolumn{4}{c|}{\textbf{KG-A2C}} & \multicolumn{4}{c|}{\textbf{RoBERTa}} & \multicolumn{4}{c}{\textbf{Swift}} \\
    \midrule
    \textbf{Task} & \textbf{$b$} & \textbf{$\alpha$} & \textbf{$m$} & \textbf{\texttt{$\alpha$} $\oplus$ \texttt{$m$}} & \textbf{$b$} & \textbf{$\alpha$} & \textbf{$m$} & \textbf{\texttt{$\alpha$} $\oplus$ \texttt{$m$}} & \textbf{$b$} & \textbf{$\alpha$} & \textbf{$m$} & \textbf{\texttt{$\alpha$} $\oplus$ \texttt{$m$}} & \textbf{$b$} & \textbf{$\alpha$} & \textbf{$m$} & \textbf{\texttt{$\alpha$} $\oplus$ \texttt{$m$}} \\
    \hline
    1 (2) &  04.82  & 04.82  & 04.82 & 04.44 & 01.48 & 01.11 & 00.74  & \textbf{03.70} & 00.74 & \textbf{01.11} & 00.00  & \textbf{01.48} & 00.00 & 00.00 & \textbf{02.22} & 00.00 \\
    % \hline
    2 (4)  &  07.07  & \textbf{07.15}  & 06.10  & 06.83 & 05.24  & \textbf{09.79}  & \textbf{13.59}  & \textbf{08.58} & 01.48  & \textbf{19.48}  & 01.25 & \textbf{01.64}  & 07.55 & \textbf{07.58} & 07.48 & 07.35  \\
    % \hline
    3  (8) &  07.67  & \textbf{09.67}  & \textbf{08.00}  & \textbf{08.33} & 07.67  & \textbf{08.33}  & 07.67  & 07.33 & 02.67  & \textbf{04.00}   & \textbf{06.67} & \textbf{05.67} & 14.60 & \textbf{16.60} & \textbf{16.60} & \textbf{51.60} \\
    % \hline
    4 (13) &  69.36  & \textbf{70.15}  & 67.70  & \textbf{69.70} & 70.96  & \textbf{72.07}  & \textbf{73.15}  & \textbf{72.15} & 67.89  & \textbf{71.11}  &  65.59 & 65.89  & 86.36 & \textbf{99.00} & \textbf{87.24} & 85.22 \\
    % \hline
    5 (16)  &  08.99  & 07.91  & \textbf{09.14}  & \textbf{09.71} & 
    05.92  & \textbf{07.15}  & \textbf{06.77}  & \textbf{06.67} & 
    08.87  & \textbf{11.48}  & 07.45 & \textbf{10.10}  & 06.39 & \textbf{07.91} & \textbf{09.70} & \textbf{08.10}  \\
    % \hline
    6 (17) &  20.34 & \textbf{22.07}  & \textbf{22.32}  & 19.94 & 22.82  & \textbf{23.07}  & 20.59  & 21.23 & 07.79  & 07.14  & 06.75 & 07.39  & 22.88 & 21.88 & 21.88 & \textbf{25.00} \\
    % \hline
    7 (21)  &  30.04  & \textbf{30.73}  & 29.69  & 28.65 & 31.42  & 29.69  & \textbf{31.94}  & \textbf{32.29} & 14.41  & \textbf{45.31} & \textbf{24.48} & \textbf{32.64} & 61.31 & \textbf{74.84} & 57.40 & \textbf{89.63}  \\
    % \hline
    8 (23) &  09.87  & 09.73  & \textbf{11.93}  & \textbf{14.07} & 15.93  & 13.53  & 15.27  & \textbf{17.53} & 04.00  & 02.67  & 02.80 & 03.42   & 45.20 & 40.20 & 12.40 & 14.00   \\
    % \hline
    9  (27) &  09.37  & 08.21  & 08.93  & \textbf{09.92} & 07.78  & \textbf{07.82}  & \textbf{08.17}  & 07.26 & 02.70  & \textbf{02.86}  & \textbf{04.68} & \textbf{02.90} & 17.14 & \textbf{17.50} & 14.40 & 15.12 \\
    % \hline
    10 (29) & 13.28  & \textbf{16.94}  & \textbf{15.13}  & \textbf{14.54} & 12.93  & 11.48  & 08.54  & 09.61 & 03.77 & 03.43 & 03.47 & \textbf{04.67}  & 17.16 & 16.24 & 17.16 & 17.11 \\
    \midrule
    \textit{avg.}  &  18.08 & \textbf{18.74} & \textbf{18.37} & \textbf{18.61} & \textbf{18.21} & \textbf{18.40} & \textbf{18.64} & \textbf{18.64} & 11.43 & \textbf{16.86} &  \textbf{12.31}  & \textbf{13.58} & 27.86 & \textbf{30.18} & 24.65 & \textbf{31.31} \\
    \bottomrule
  \end{tabular}
  % }
  % \vspace{-0.3cm}
\end{table*}

\section{Experimental Setup}

\subsection{Task and Evaluation Metrics}
\task~is a virtual representation of the world in an intricate text-based environment in English, with a variety of objects, actions, and tasks \cite{s1}. 
It includes ten connected locations with 218 unique objects such as instruments, electrical components, plants/animals, substances, containers, and everyday objects like furniture, books, and paintings. 
There are 25 high-level actions, with up to 200,000 possible combinations per step, only a few of which have practical applications. \task~has 10 tasks with a total set of 30 sub-tasks. 
Due to the diversity within \task, each task functions as an individual benchmark with distinct reasoning abilities, knowledge requirements, and varying numbers of actions needed to achieve the goal state.    
Moreover, each sub-task has a set of mandatory objectives that need to be met by any agent (such as focusing on a non-living object and putting it in a red box in the kitchen). 
The rewards for completing these tasks are highly quantized for learning purposes to guide the agent toward preferred solutions. Namely, for each performed action, the \task~environment provides a numeric score (reward) and a boolean indication of whether the task has been completed. The agent can take up to 100 steps (actions) in each episode, and its final score is scaled to fall between 0 and 100. Its score improves when both the episode goal and its sub-goals are achieved. The evaluation for an episode concludes and the cumulative score is returned when the agent receives information from the environment that the task has been completed or the limit of 100 steps is reached. %, then the evaluation concludes .

For experimentation purposes, we selected a single representative sub-task from each of the 10 tasks. %, which serve as representative examples.
The numbers in brackets in the `Task' column of Table \ref{tab:best} signify the original \task~sub-task number out of 30.\footnote{Please refer to \cite{s1} for more information about the tasks and their train-test splits.}
% \subsection{Evaluation Metrics}
All evaluation results in this paper are averaged over three model runs on the test dataset. %We report the average of these three agent scores. % is reported. 

\subsection{Implementation and Modeling Details}

% \textbf{Model Training.} 
Following the original methods, we use task-specific training for DRRN, KG-A2C, and RoBERTa, resulting in the creation of 10 distinct models for the 10 tasks. In contrast, Swift is trained once using the entire training dataset. While we conducted experiments with KG-A2C and RoBERTa to develop a unified model for a more fair comparison, the outcomes were detrimental to the performance. Hence, we use the same setup of DRRN and KG-A2C as in \task.
% \noindent \textbf{Hyperparameters.} 
DRRN is trained with a learning rate of $1e^{-4}$, an embedding dimension of 128, and a hidden dimension of 128. KG-A2C uses a learning rate of $3e^{-3}$, a dropout rate of 0.2, an embedding dimension of 50, and a hidden dimension of 100. DRRN and KG-A2C utilized eight parallel environments to speed up the training process. 
These parameter values have been taken from the original \task~paper.
For RL models we perform training for 40,000 steps as we were able to reproduce the baseline results with 5\% of the original \task~paper.
For the RoBERTa model, we use \textit{roberta-large} for all of the experiments. For training, we use 3 epochs,\footnote{A common choice across NLP tasks, further tuning did not yield improvements.} a learning rate of $2e^{-5}$, 4 randomly selected distractors, and a batch size of 1. For RoBERTa's MCA variants, we use a window size of $A=5$. 
For Swift, we use \textit{flan-T5-base} with a learning rate of $1e^{-4}$ and a batch size of 6. The maximum source and target lengths are set to 1,024 and 16, respectively.
For Swift model, we used 8 training epochs following the original paper.

\noindent \textbf{Environment.} We used two identical servers, each with an Intel(R) Xeon(R) Gold 5215 CPU @ 2.50GHz, featuring 40 cores and 256 GB of RAM. We also utilized eight NVIDIA RTX A5000 GPUs (per server) to accelerate the training and inference process.

\section{Results \& Analysis}

\subsection{Effect of Knowledge Injection}
\textbf{Overall results.} Table \ref{tab:best} compares our best model with \texttt{baseline}: in 34 out of 40 cases, our knowledge injection strategies improve over the baseline models. Among these cases, the most successful strategy is including affordances, which obtains the best results in 15 cases, followed by including MCA (8 cases). Including both knowledge types together led to the best results in 11 cases. 
The positive effect of adding affordances is confirmed in Table \ref{tab:affordance_pca}, which shows that including affordances improves the selection of the subsequent best action in 63\% (25 out of 40) of cases. While the integration of affordances has a positive overall impact on the agents' action selection, in another 13 cases including affordances harms the model performance. Including the memory of previous correct actions taken by the agent also effectively enhances the decision-making capabilities of the architectures under consideration, though to a lesser extent compared to including affordances (Table \ref{tab:affordance_pca}).
Given the varying
effectiveness of affordances and MCA, 
% to architectural factors and task-related factors such as the complexity of the game and the size of the action space. W
we next study the performance variations across models and tasks.

\begin{figure*}[!t]
    \centering
    \includegraphics[width=0.67\linewidth]{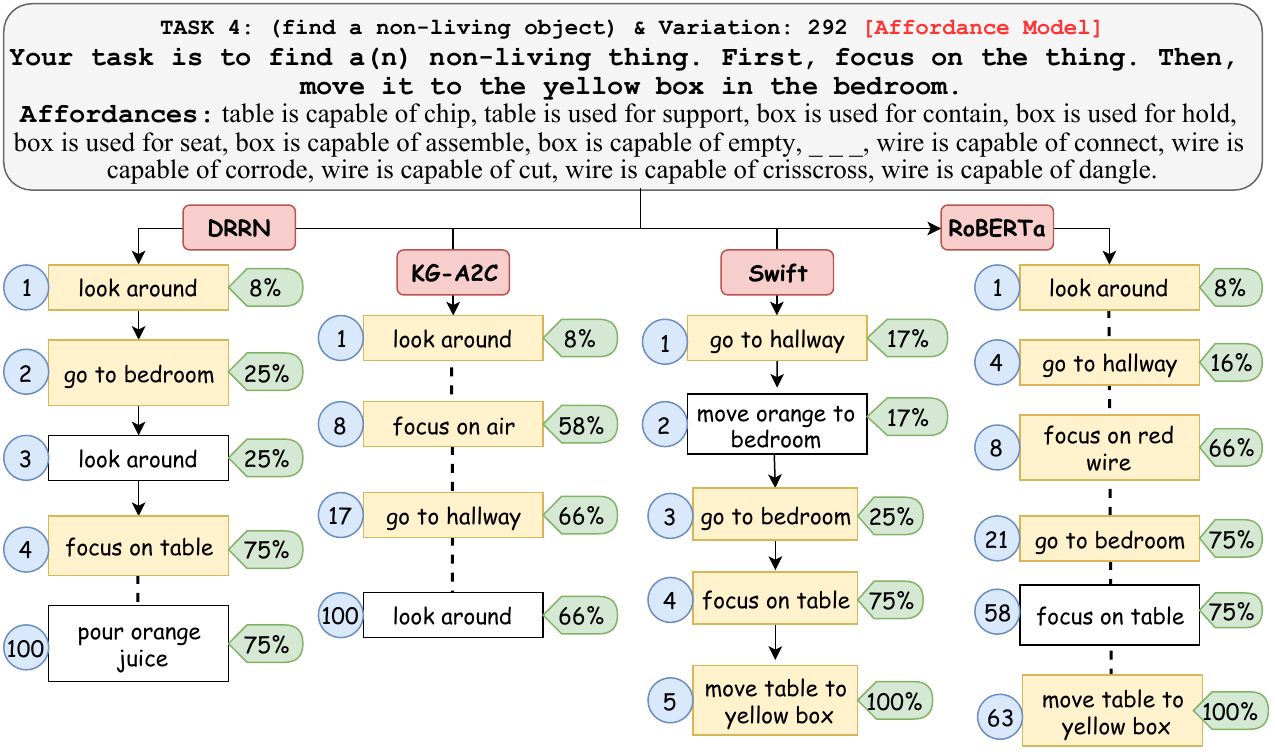}
     % \vspace{-0.2cm}
    \caption{Actions taken by affordance models on Task 4. Blue = step index, green = cumulative score, and yellow = correct action.}% taken by the agent.}
    \label{appendix1}
    % \vspace{-0.3cm}
\end{figure*}

\textbf{Performance variations across architectures.}
To study further the isolated effect of different types of injected knowledge, we compare the model performance with and without knowledge injection for the four models.
The RL-based DRRN model benefits from affordances most consistently, with 9/10 tasks showing the best performance after affordances are included (Table \ref{tab:best}), leading to a 4\% relative increase in performance (Table~\ref{tab:affordance_pca}). The DRRN model relies on exploring the action space to learn optimal policies, and providing affordance information allowed the model to narrow down the search space and focus on actions that lead to successful outcomes.
In terms of the overall impact across tasks, the LM variants, RoBERTa and Swift, benefit the most on average from including affordances, leading to a relative increase of 48\% and 8\% respectively, over the baselines. Affordances improve the score of the KG-A2C method in 6/10 cases, yet, the overall improvement over this baseline is marginal. 
For DRRN and KG-A2C, in slightly over half of the cases, integrating MCA improves performance in selecting the next best action by 2\% relative to the baseline. Interestingly, MCA improves over the RoBERTa baseline by approximately 8\% in relative terms, despite only helping in 3/10 tasks. 
Furthermore, eliminating the action history proves advantageous for Swift.

\textbf{Performance variations across tasks.} Table \ref{tab:affordance_pca} shows granular performance per task for all four models with their corresponding knowledge-injection variants. While we note that the impact of knowledge varies across tasks, in most cases, the performance is boosted by either of the knowledge-injection strategies.
We note that task 3 (\textit{Electricity}) is the only one where both knowledge injection strategies help across all architectures. Here,
the DRRN and KG-A2C models experience an average increase of around 10\% (relative) in performance, while the RoBERTa and Swift models show an average of 100\% and 14\% relative increase in performance.
An example goal in task 3 is to power a red light bulb using a renewable power source, which requires that the agents understand the affordances of the electrical circuitry involved and the renewable energy sources that can be used to power it. The affordances provide the agent with valuable information that the light bulb is capable of generating light. Furthermore, the agent acquired the ability to remember its prior successful selection of a light bulb, which facilitated the subsequent selection of the wire and solar panel while avoiding the repetition of its prior choice.

Meanwhile, we observe that tasks 8 and 10 require biological knowledge, while the affordances retrieved from ConceptNet contain information like `dog \textit{capableOf} \{bark, guard\}' %and `dog \textit{capableOf} guard' 
that are not informative for inferring the lifespan or life stages of a dog.
Alternatively, the addition of affordance significantly improves the performance of the RoBERTa model in tasks 2 and 7, leading to 13x and 3x better performance, respectively. Moreover, RoBERTa with affordances achieves perfect scores 14 and 9 times for tasks 4 and 7, respectively, which is rare, especially given the relatively large sequences of correct actions in the \task~tasks. Notably, tasks 4 and 7 have an average length of less than 10, indicating that the model performs well in shorter tasks. 
The highest improvement for RoBERTa happens on task 7, which has the shortest sequence of correct actions on average. 
Swift ($\alpha \oplus m$) experienced a substantial performance improvement (with a 3.5x increase over the baseline) on task 3, where the agent successfully achieved the goal state in 40\% of the episodes. Moreover, in 96\% of the cases for task 4, the affordance variant of Swift was able to get a perfect score of 100\%.
These results strongly suggest the application of techniques that tailor the learning process to the specific task at hand, like meta-learning~\cite{huisman2021survey}, to empower the system to intelligently discern and apply the most suitable knowledge for optimal performance and adaptability.

\subsection{Effect of Affordances}

Given that affordances are a more effective knowledge-injection strategy than including the MCA, we perform a case study of injecting affordances in different models and we compare ways to inject affordances into KG-A2C.

\textbf{Case study.}
Figure \ref{appendix1} presents a case study regarding the models' ability to incorporate affordance information for task 4. We opted for this task of finding a non-living object given the relatively high performance of all models on it.
We see that the affordance `\textit{wire} capable to connect' enhanced RoBERTa's comprehension of wires as non-living objects, yielding a positive environment reward at step 8.
The LMs, as well as DRRN, also utilized the affordances associated with the \textit{table} object (e.g., `table is capable of support') to identify it as a non-living object. 
The affordances associated with the term \textit{box} (such as \textit{`box is used for contain'} and \textit{`box is used for hold'}) enhanced the LMs' grasp of the box's attributes, facilitating the execution of the final action. 
While both LMs benefited from the affordance knowledge, RoBERTa required 63 steps to finish the episode, while Swift completed the task in just five steps. 
This supports our experimental finding that, compared to Swift, RoBERTa takes more time to pick the correct action.
The RL agents (DRRN and KG-A2C) faced challenges in achieving perfect scores within this sub-task. KG-A2C struggled to reach the intended destination (the bedroom), often navigating to other locations and performing arbitrary actions. While DRRN managed to reach the bedroom and obtained a slightly better score, it encountered difficulty locating the box despite the provision of affordances.
This case study suggests that LMs such as RoBERTa and Swift apply affordance knowledge more effectively than RL methods for such tasks.

\begin{figure}[!t]
    \centering
    \includegraphics[width=\linewidth]{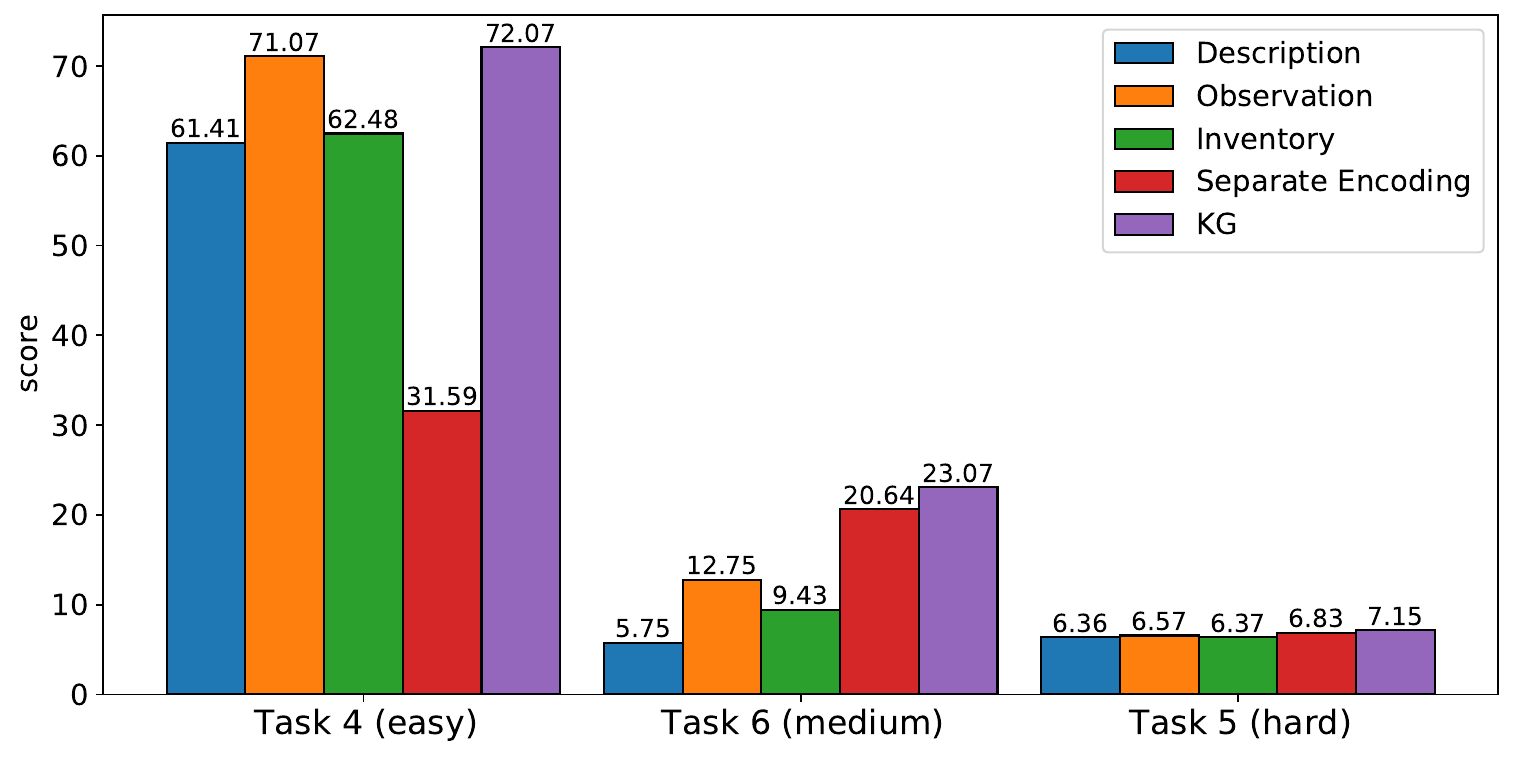}
    % \vspace{-0.4cm}
    \caption{Effect of five ways to add affordances in KG-A2C.}
    % \vspace{-0.5cm}
    \label{affordance_comparison}
\end{figure}

\textbf{Optimal way to inject affordances.} We have chosen KG-A2C to conduct the ablation study, as it has a larger number of modular components (KG, graph attention, and actor-critic module), which can be flexibly manipulated for experimentation. Moreover, KG-A2C benefits the least from affordance injection.
We explore multiple variations of injecting affordance knowledge into KG-A2C: by adding it as input into the observation, inventory, and description, creating a separate GRU encoding layer for affordance, and adding affordance to the KG itself. 
We evaluate the performance of each method on three sub-tasks: easy (task 4), medium (task 6), and hard (task 5), based on the number of actions and the performance of the baseline models. The results in Figure \ref{affordance_comparison} consistently suggest that the incorporation of affordances as part of the KG performs better than including them as part of the other components (e.g., description) or encoding them separately. 
A possible explanation is that by adding affordances to the KG, we allow the agent to have a more structured and separate representation of the environment, which in turn helps the agent make more informed decisions. Adding affordances as strings concatenated to inputs or adding a separate encoding layer hurts performance; we think that these methods cause information overload or interference with the original inputs, thus confusing the agent. The separate encoding layer introduces additional complexity to the architecture, making it harder for the agent to learn and generalize, especially considering the limited data size. Meanwhile, we note that an alternative approach to incorporate affordances is via self-supervision via auxiliary tasks, which brings significant improvement for some tasks in the case of RoBERTa, and suggests an avenue for RL-LM integration.

\section{Conclusions and Outlook}
This paper investigated whether current AI agents can use knowledge injection in semi-Markov text-based games to act coherently and improve their ability to learn from the environment through enhanced contextual awareness. We proposed to inject knowledge about affordances and keep a memory of previous correct actions on diverse architectures.
Through rigorous evaluation, we showed improvement over the four baseline models across ten elementary school tasks. Among our injection methods, affordance knowledge was more beneficial than the memory of correct actions. The variable effect across tasks was frequently due to the relevance of the injected knowledge to the task at hand, with certain tasks (e.g., task 3: electricity) benefiting more from the injection. 
Injecting affordances was most effective via KGs; incorporating them as raw inputs increased the learning complexity for the models.
The insights into the usage of knowledge injection for improving the performance of RL and LMs in complex IF games have potential implications for interactive applications beyond the gaming domain, including customer service chatbots and personal assistants.

As the resulting models' performance is still far from ideal,
we envision two future directions toward more coherent and efficient models.
First, our results suggest that the models have complementary strengths and weaknesses: the RL model performed the best on the task \textit{Matter} (task 1), the KG-augmented model yielded the best performance on the task of \textit{Measurement} (task 2), and the language models outperformed the others on \textit{Biology I} (task 5), \textit{Biology II} (task 7), and \textit{Biology IV} (task 10). Inspired by this insight, we propose to enhance the performance of the LMs by incorporating an RL policy network~\cite{ziegler2019fine}. 
Second, few-shot prompting of large LMs has recently shown promise on reasoning tasks, as well as clear benefits from interactive communication and input clarification~\cite{madaan2022memory}. Exploring their role in interactive tasks, either as solutions that require less training data or as components that can generate synthetic data for knowledge distillation to smaller models, is another promising future direction.

\bibliographystyle{ACM-Reference-Format}
\bibliography{sample-base}

\clearpage
\onecolumn
\appendix
\section{Additional Task Details}

\subsection{Task Descriptions}
\begin{enumerate}
    \item \textbf{Task 1 - \textit{Matter}:} Your task is to freeze water. First, focus on the substance. Then, take actions that will cause it to change its state of matter.
    \item \textbf{Task 2 - \textit{Measurement}:} Your task is to measure the melting point of chocolate, which is located around the kitchen. First, focus on the thermometer. Next, focus on the chocolate. If the melting point of chocolate is above -10.0 degrees, focus on the blue box. If the melting point of chocolate is below -10.0 degrees, focus on the orange box. The boxes are located around the kitchen.
    \item \textbf{Task 3 - \textit{Electricity}:} Your task is to turn on the red light bulb by powering it using a renewable power source. First, focus on the red light bulb. Then, create an electrical circuit that powers it on.
    \item \textbf{Task 4 - \textit{Classification}:} Your task is to find a(n) non-living thing. First, focus on the thing. Then, move it to the red box in the kitchen.
    \item \textbf{Task 5 - \textit{Biology I}:} Your task is to grow a apple plant from seed. Seeds can be found in the kitchen. First, focus on a seed. Then, make changes to the environment that grow the plant until it reaches the reproduction life stage.
    \item \textbf{Task 6 - \textit{Chemistry}:} Your task is to use chemistry to create the substance 'salt water'. A recipe and some of the ingredients might be found near the kitchen. When you are done, focus on the salt water.
    \item \textbf{Task 7 - \textit{Biology II}:} Your task is to find the animal with the longest life span, then the shortest life span. First, focus on the animal with the longest life span. Then, focus on the animal with the shortest life span. The animals are in the 'outside' location.
    \item \textbf{Task 8 - \textit{Biology III}:} Your task is to focus on the 4 life stages of the turtle, starting from earliest to latest.
    \item \textbf{Task 9 - \textit{Forces}:} Your task is to determine which of the two inclined planes (unknown material C, unknown material H) has the most friction. After completing your experiment, focus on the inclined plane with the most friction.
    \item \textbf{Task 10 - \textit{Biology IV}:} Your task is to determine whether blue seed color is a dominant or recessive trait in the unknown E plant. If the trait is dominant, focus on the red box. If the trait is recessive, focus on the green box.
\end{enumerate}

\subsection{ScienceWorld Gameplay Example}

% \subsubsection{DRRN}
\textbf{Task:} 4 (find a non-living thing)\\
\textbf{Variation:} 239 (DRRN baseline)\\
\textbf{Description:} Your task is to find a(n) non-living thing. First, focus on the thing. Then, move it to the purple box in the workshop.\\

\footnotesize{
\noindent  \texttt{\noindent \textbf{Action 1:} look around\\
\textbf{Freelook:} This room is called the workshop. In it, you see: - a ultra low temperature freezer. The ultra low temperature freezer door is closed. - a substance called air - the agent - a table. On the table is: a red light bulb, which is off, a red wire, a orange wire, a yellow light bulb, which is off, a green wire, a switch, which is off, a violet light bulb, which is off, a battery. - a purple box (containing nothing). You also see: - A door to the hallway (that is open)\\ 
\textbf{Inventory:} In your inventory, you see: an orange\\
\textbf{Observation:} This room is called the workshop. In it, you see: - a ultra low temperature freezer. The ultra low temperature freezer door is closed. - a substance called air - the agent - a table. On the table is: a red light bulb, which is off, a red wire, a orange wire, a yellow light bulb, which is off, a green wire, a switch, which is off, a violet light bulb, which is off, a battery. - a purple box (containing nothing). You also see: A door to the hallway (that is open)\\
\textbf{Score:} 16.66\% \\ \\ 
\textbf{Action 2:} pour red light bulb into hallway\\
\textbf{Freelook:} This room is called the workshop. In it, you see:  - a ultra low temperature freezer. The ultra low temperature freezer door is closed. - a substance called air - the agent - a table. On the table is: a red light bulb, which is off, a red wire, a orange wire, a yellow light bulb, which is off, a green wire, a switch, which is off, a violet light bulb, which is off, a battery. - a purple box (containing nothing). You also see: A door to the hallway (that is open) \\
\textbf{Inventory:} In your inventory, you see: an orange\\
\textbf{Observation:} You pour the red light bulb into the hallway.\\
\textbf{Score:} 16.66\% \\ \\ 
\textbf{Action 3:} look at inventory\\
\textbf{Freelook:} This room is called the workshop. In it, you see: - a ultra low temperature freezer. The ultra low temperature freezer door is closed. - a substance called air - the agent - a table. On the table is: a red wire, a orange wire, a yellow light bulb, which is off, a green wire, a switch, which is off, a violet light bulb, which is off, a battery. - a purple box (containing nothing). You also see: A door to the hallway (that is open)\\
\textbf{Inventory:} In your inventory, you see: an orange\\
\textbf{Observation:} a inventory (containing an orange)\\
\textbf{Score:} 16.66\% \\ \\ 
\textbf{Action 4:} move battery to table\\
\textbf{Freelook:} This room is called the workshop. In it, you see: - a ultra low temperature freezer. The ultra low temperature freezer door is closed. - a substance called air - the agent - a table. On the table is: a red wire, a orange wire, a yellow light bulb, which is off, a green wire, a switch, which is off, a violet light bulb, which is off, a battery. - a purple box (containing nothing). You also see:  door to the hallway (that is open)\\
\textbf{Inventory:} In your inventory, you see: an orange\\
\textbf{Observation:} You move the battery to the table.\\
\textbf{Score:} 16.66\% \\ \\ 
\textbf{Action 5:} focus on switch \\
\textbf{Freelook:} This room is called the workshop. In it, you see:  - a ultra low temperature freezer. The ultra low temperature freezer door is closed.  - a substance called air - the agent - a table. On the table is: a red wire, a orange wire, a yellow light bulb, which is off, a green wire, a switch, which is off, a violet light bulb, which is off, a battery. - a purple box (containing nothing). You also see: A door to the hallway (that is open)\\
\textbf{Inventory:} In your inventory, you see: - an orange\\
\textbf{Observation:} You focus on the switch\\
\textbf{Score:} 66.66\% \\ \\ 
\textbf{Action 6:} look at anode in violet light bulb\\
\textbf{Freelook:} This room is called the workshop. In it, you see:  - a ultra low temperature freezer. The ultra low temperature freezer door is closed.  - a substance called air - the agent - a table. On the table is: a red wire, a orange wire, a yellow light bulb, which is off, a green wire, a switch, which is off, a violet light bulb, which is off, a battery. - a purple box (containing nothing). You also see: A door to the hallway (that is open)\\
\textbf{Inventory:} In your inventory, you see: - an orange\\
\textbf{Observation:} a anode. it is connected to: nothing\\
\textbf{Score:} 66.66\% \\ \\ 
\textbf{Action 7:} move table to purple box\\
\textbf{Freelook:} This room is called the workshop. In it, you see:  - a ultra low temperature freezer. The ultra low temperature freezer door is closed.  - a substance called air - the agent - a table. On the table is: a red wire, a orange wire, a yellow light bulb, which is off, a green wire, a switch, which is off, a violet light bulb, which is off, a battery. - a purple box (containing nothing). You also see:  A door to the hallway (that is open)\\
\textbf{Inventory:} In your inventory, you see: - an orange\\
\textbf{Observation:} You move the table to the purple box\\
\textbf{Score:} 100\%
}
}

\end{document}